\documentclass{article}
\usepackage[preprint,nonatbib]{neurips_2021}
\usepackage[utf8]{inputenc} % allow utf-8 input
\usepackage[T1]{fontenc}    % use 8-bit T1 fonts
\usepackage{url}            % simple URL typesetting
\usepackage{booktabs}       % professional-quality tables
\usepackage{amsfonts}       % blackboard math symbols
\usepackage{nicefrac}       % compact symbols for 1/2, etc.
\usepackage{microtype}      % microtypography
\usepackage{xcolor}         % colors

\usepackage{dirtytalk}
\usepackage{hyperref}

\title{Improving Generalization in Mountain Car Through the Partitioned Parameterized Policy Approach via Quasi-Stochastic Gradient Descent}
\author{%
  Caleb M. Bowyer \\
  Department of Electrical and Computer Engineering\\
  University of Florida\\
  Gainesville FL, USA \\
  \texttt{c.bowyer@ufl.edu} \\
}

\usepackage{cite}
\usepackage{amsmath,amssymb,amsfonts}
\usepackage{algorithmic}
\usepackage{graphicx}
\usepackage{textcomp}
\usepackage{xcolor}

\begin{document}

\maketitle

\begin{abstract}
% The reinforcement learning problem of finding a control policy that minimizes the minimum time objective for the Mountain Car environment is considered. Particularly, a class of parameterized nonlinear feedback policies is optimized over to reach the top of the highest mountain peak in minimum time. The optimization is carried out using quasi-Stochastic Gradient Descent (qSGD) methods. In attempting to find the optimal minimum time policy, a new parameterized policy approach is considered that seeks to learn an optimal policy parameter for different regions of the state space, rather than rely on a single macroscopic policy parameter for the entire state space. This partitioned parameterized policy approach is shown to significantly outperform the uniform parameterized policy approach. 

The reinforcement learning problem of finding a control policy that minimizes the minimum time objective for the Mountain Car environment is considered. Particularly, a class of parameterized nonlinear feedback policies is optimized over to reach the top of the highest mountain peak in minimum time. The optimization is carried out using quasi-Stochastic Gradient Descent (qSGD) methods. In attempting to find the optimal minimum time policy, a new parameterized policy approach is considered that seeks to learn an optimal policy parameter for different regions of the state space, rather than rely on a single macroscopic policy parameter for the entire state space. This partitioned parameterized policy approach is shown to outperform the uniform parameterized policy approach and lead to greater generalization than prior methods, where the Mountain Car became trapped in circular trajectories in the state space. 
\end{abstract}

\section{Background and Literature Overview}
Generalization in reinforcement learning problems is a difficult problem. This paper aims to solve the generalization problem more broadly for continuous space RL tasks, but analyzes a new method called the partitioned parameterized policy approach on a particular problem, the Mountain Car environment to validate the methodology. In this environment it has been researched and found that after training various algorithms for Mountain Car the following effect was observed: \say{the car has oscillated back and forth in the valley, following circular trajectories in state space.} \cite{sutton2018reinforcement}. Furthermore, it is shown through simulation that the new method that involves learning policy parameters for distinct regions of the state space improves the performance that is possible by a single policy parameter. The continuous state and control space Mountain Car example for RL has its origins in the dissertation \cite{moore1990efficient}. This control example has appeared as a benchmark problem in several different RL problem settings. Some of these different RL methods tested include \cite{suttongeneralization}, \cite{singh1996reinforcement}, and a setting with noisy observations in \cite{heidrich2008variable}. The issues of applying RL in the continuous space setting was made clear in the article \cite{smartpractical}. The form of the discrete-time mountain car RL example considered in this paper most closely resembles the example as featured in the popular RL textbook \cite{sutton2018reinforcement}, and the recent paper \cite{melnikov2014projective}. The next section clearly presents the Mountain Car environment and the minimum time objective considered in this paper. 
%%%%%%%%%%%%%%%%%%%%%%%%%%%%%%%%%%%%%%%%%%%%%%%%%%%%%%%%%%%%%%%%%%%%%%%%%%%%%%%%
\section{Discrete-Time Mountain Car Environment and Minimum-Time Objective Function}
Based on forces acting on a car driving up a slope or along a road it is not difficult to derive the state-space equations for the two dimensional continuous-time system. I first present the continuous-time model before presenting the discrete-time model used in the rest of the experiments in this paper. The state space is two dimensional for Mountain Car. Denote the position in continuous time as $z_t = x_1(t)$ and the velocity as $v_t = x_2(t)$. The state is represented as $$x_t=[z_t,v_t]^T\in \mathcal{X}=[z_{min}, z_{goal}]^T\times[v_{min},v_{max}]^T.$$ The Mountain Car's state-space model is given by:

\begin{gather*}\label{mountain_car_cts}
% \begin{equation} \label{obs_eqn}
\begin{split} 
            \begin{bmatrix}
                \frac{dx_1}{dt} \\
                \\
                \frac{dx_2}{dt}
                \end{bmatrix}
                = \begin{bmatrix}
                x_2\\
                \\
                \frac{k}{m}u - gsin(\theta(x_1))
                \end{bmatrix}.
\end{split}
%\end{equation}
\end{gather*}

The parameters $k,m,g$ are constants and set according to \cite{sutton2018reinforcement}. Some practical restrictions to the phase-space trajectories of the Mountain Car follow from the bounded continuous-state space of the problem. Depending on how $x_1$ is \say{outside} of the interval $[z_{min}, z_{goal}]$, different effects arise in the simulation of a control policy. The state $z_{goal}$ is the maximum displacement to the right and also denoted by $z_{max}$. If the car enters this part of the state space for any velocity, it is absorbing in the sense that the run is cut short and no further cost is incurred. The finalized discrete-state equation for the Mountain Car problem is described next. A simplification for the angle $\theta$ is made so that $\theta(x_1) = \pi + 3x_1$, $\frac{k}{m} = 10^{-3}$, and $g=2.5\times10^{-3}$. The specific parameter values that were selected come from Sutton's textbook description of the problem \cite{sutton2018reinforcement}. Then, the discrete-time state space model reduces accordingly:

\begin{gather*}\label{mountain_car_discrete}
% \begin{equation} \label{obs_eqn}
\begin{split} 
            \begin{bmatrix}
                z_{k+1} \\
                \\
                v_{k+1}
                \end{bmatrix}
                = \begin{bmatrix}
                z_k + v_k\\
                \\
                v_k + 10^{-3}u_k -2.5\times10^{-3} cos(3z_k)
                \end{bmatrix},
\end{split}
%\end{equation}
\end{gather*}
where the continuous valued states are projected to be inside of the bounded state-space box as described previously. One other case must be mentioned before the size of the box is made explicit, also determined in \cite{sutton2018reinforcement}. For any negative velocity if the car ever reaches $z_{min}$, then the velocity is reset to $v_k = 0$ as the physics would dictate in real life for a car crashing into the leftmost wall or obstacle in the environment. The intervals for the states are now fixed as follows:

$$ [z_{min}, z_{goal}] = [-1.2,0.5], $$ and
$$ [v_{min},v_{max}] = [-7\times10^{-2}, 7\times10^{-2}].$$ These values and constants are fixed for the rest of the experiments. Before discussing the parameterized policy approaches and how they are optimized with qSGD, I first describe the minimum time objective for this RL problem. One of the primary goals of this paper is to learn a minimum time policy. Hence, our cost-to-go function must reflect that goal. For this goal, a fixed maximum time-horizon is set $T_{max}$; however, the run or episode could be shorter depending on whether the car reaches the goal state or not. If the car reaches the goal state at some time $T < T_{max}$, then the episode is $T$ steps long. It is shown later that most of the runs end up in the interval $[T_{low}, T_{high}] = [40, 48]$. This level of performance is more than an order of magnitude better than the performance achieved in \cite{sutton2018reinforcement}, hinting at the superiority of qSGD methods as compared to the SARSA($\lambda$) algorithms considered there for continuous state RL control problems. All of the episodes using the SARSA($\lambda$) algorithms required steps in the interval $[400, 700]$ long. The stated RL objective of minimum time to reach the goal state from anywhere leads to the following cost-to-go form:

$$ J^*(x) = \min\limits_{\bf{u}}\sum\limits_{k=0}^{T_{max}} c(x_k, u_k),$$ 

where $x_0=x$, and $\bf{u}$ is the sequence of controls until the end of the run. To enforce the stated objective of reaching the top of the hill in minimum time, the cost is set to $c(x,u)=1~\forall x,u$ with $z \neq z_{goal}$, i.e., for all non-goal states. Zero cost is only obtained at the absorbing goal state, i.e. $c([z_{goal},v]^T,u) = 0~\forall u,v$. 

Denote the cost-to-go based on some policy parameter $\theta$ from some state $x$ as $J^{\theta}(x)$. Then, for $K$ initial conditions on the state, one can consider the cost signal to be used in the qSGD algorithm as:

$$\Gamma^{\bullet}(\theta) = \frac{1}{K}\sum\limits_{i=1}^K J^{\theta}(x_0^{i}),$$
where $\{x_0^i\}$ are the set of pre-selected initial conditions on the state used for training the parameters of every policy that is learned. Without enforcing a timeout through $T_{max}$, almost every $\theta$ would result in $\Gamma^{\bullet}(\theta) = \infty$. The final tweak to the cost measure is $\Gamma(\theta) = min(\Gamma^{\bullet}(\theta),~T_{max})$, which is the minimum time objective based on our policy parameter $\theta$. The policy parameter (parameters) $\{\theta\}$ is learned to minimize this objective function using qSGD described briefly in the next section for the uniform (partitioned) policy approach. A set of testing initial conditions are also saved and used to compare the generalizability of the different parameterized policy approaches that are analyzed, to test the algorithms under non-ideal conditions.

% A set of training intial conditions will be maintained for the policy parameters as well as a set of testing initial conditions to also test under non-ideal conditions.

\section{Analysis and Motivation of parameterized policy family}
The parameterized class of policies is based on controlling the total energy of the system so as to maximize the potential energy for reaching the top of the hill. Start by considering the kinetic energy (KE) and potential energy (PE) of the car. The KE is given by $KE(x_2) = \frac{1}{2}mx_2^2$, and the PE is given by $PE(x_1) = \frac{1}{3}mgsin(x_1)$. The goal is to maximize the total energy $E = KE+PE$ to reach the mountain top. I design $\phi^{\theta}$ to minimize the negative of the total energy to cast the problem as a minimization. Denote the Lyapunov function by $J(x) = \frac{1}{2}E^2$. The time derivative yields:

\begin{align*}
\dot{J}(x) &= E\dot{E}\\
            &=E \frac{d}{dt}\biggl[\frac{1}{2}m x_2^2 + \frac{1}{3}mgsin(x_1)\biggr]\\
            &=E\biggl[mx_2\dot{x}_2+\frac{1}{3}mgcos(x_1)\dot{x}_1\biggr].
\end{align*}
After substituting the dynamics and simplifying one obtains:
\begin{align*}
\dot{J}(x) &= kx_2 E u + mgEsin(\pi+3x_1)x_2 + \frac{1}{3}mg E cos(x_1)x_2\\
           &= mg x_2 E\biggl[sin(\pi+3x_1) + \frac{cos(x_1)}{3}\biggr] + kx_2Eu.
\end{align*}

The form of the feedback law is based on minimizing over controls the following equation:

$$\phi_J(x) = \arg\min\limits_u \biggl\{\frac{1}{2}Ru^2 + f(x,u)\cdot\nabla J(x)\biggr\}.$$ 

Taking the derivative of the expression with respect to $u$ and setting equal to zero yields the feedback control $u = -\frac{k}{R}x_2 E$, which is to be applied to reach the goal state asymptotically. Simplifying further, the final form of the nonlinear feedback policy is more clearly seen:

$$u = -\frac{km}{2R}x_2^3 - \frac{kmg}{3R}x_2 sin(x_1).$$ This control solution is an asymptotically stable solution, but it does not guarantee that the state converges to the goal state in minimum time. To ensure converge to the goal state and do so in minimum time, I use the following parameterized family based on the nonlinear feedback, and optimize over $\theta$ to converge to the minimum time policy for this class of converging policies. Furthermore, since this is a constrained optimization problem ($-1\leq u \leq 1$), this leads to the following form for the parameterized policy:

% The class of parameterized policies takes the following form:
$$\phi^{\theta}(x) = \text{sign}\biggl\{\theta_1 x_2^3 + \theta_2x_2sin(x_1)\biggr\}.$$

The parameter $\theta = [\theta_1, \theta_2]^T$ is optimized to minimize the minimum time objective $\Gamma(\theta)$ previously introduced. This form of parameterized nonlinear policy is experimented with using both a uniform approach and a partitioned approach in the simulation section. I optimize the parameter $\theta$ using the qSGD $\#1$ algorithm from the RL textbook (forthcoming) \cite{CSRL}. The next section introduces and explains the qSGD algorithm.

\section{The qSGD Algorithm}
The qSGD algorithm utilizes the parameter ODE approach to RL and quasi-stochastic approximation (qSA) where exploration is driven by deterministic sinusoids or other oscillatory signals in time \cite{CSRL}. The more classical approaches to exploration have relied on identically independent distributed (iid) samples for exploration, but this is much slower in the convergence of the parameter ODE that solves a special root finding problem. These qSA methods have been applied in gradient free optimization and policy gradient RL algorithms. The qSA method involves solving a root finding problem of the following form:

$$\Bar{f}(\theta) = E[f(\theta,\xi)],$$
where $\theta\in \mathbb{R}^d$ is the parameter that is learned through the qSGD algorithm to be explained. The variable $\xi$ is the deterministic noise, which for this paper is a sum of two sinusoids at different frequencies and amplitudes, i.e., $\xi\in \mathbb{R}^2$. The vector $\Bar{f}$ plays an analogous role as the vector field $f$ that defines an ODE. The end goal of the qSGD algorithm is to find a globally optimal $\theta$ if possible so that $\Bar{f}(\theta^*) = 0$, for some $\theta^*$ in the parameter space. A parameter ODE in $\theta$ is created in continuous time using  $\Bar{f}$ and translated into a discrete time algorithm, most often using the standard Euler approximation. Other translations are possible: \cite{Meyn2000ode}, \cite{bowyer2021predictor}, and \cite{Devraj}. A menu of algorithms is presented in \cite{CSRL}, where the signal $\Psi_t = \theta_t + \epsilon\xi_t $, for $\epsilon>0$, and $\xi_t$ is the $d$ dimensional probing signal. This deterministic exploration or probing signal is preferred because it has been shown to lead to faster convergence than the iid type of exploration for parameter convergence \cite{CSRL}. In this paper, $d=2$ for both the uniform and partitioned policy that is to be analyzed. The qSGD $\# 1$ algorithm is presented as such from \cite{CSRL}:

\begin{quote}
    For a given $d \times d$ positive definite matrix $G$, and $\theta_0\in \mathbb{R}^d$,
    $$\frac{d}{dt}\theta_t = -a_t\frac{1}{\epsilon} G \xi_t \Gamma(\Psi_t),$$ and
    $$ \Psi_t = \theta_t + \epsilon \xi_t.$$
\end{quote}

Applying the Euler approximation yields an algorithm of the form:

$$\theta_{n+1} = \theta_n + a_{n+1}\frac{1}{\epsilon} G \xi_{n+1}\Gamma(\Psi_{n+1}),$$ and
$$\Psi_{n+1} = \theta_n + \epsilon\xi_{n+1}.$$

Some final simplifications are made to the discrete time RL algorithm next. Now, I discuss my probing signal and other constants that are used in the rest of the experiments. A very general probing signal for a single dimension may be constructed as:
$$\xi_t = \sum\limits_{i=1}^{\ell} a^i sin(2\pi(\omega^it+\phi^i)),$$ where each sinusoid in the mixture could have a different amplitude, frequency and phase. One option for setting these parameters is to randomly sample them from an interval after $\ell$ is fixed, i.e., the number of sinusoids in the mixture. For my experiments $\ell=2,~d=2$ and the phase is set to zero. Furthermore, the amplitudes are randomly sampled as $a^1,a^2 \sim \text{unif}((0,1))$. The frequencies were selected to have one slowly varying and one varying much faster by setting $w^1= 0.3, w^2=50$. The algorithm was further simplified by setting $G=I$, i.e. to the identity matrix, $\epsilon=1$. The final discrete-time algorithm took the following form:

$$\theta_{n+1} = \theta_{n} - \frac{g}{(1+n)^{\rho}}\xi_n \Gamma(\Psi_n),$$

$$\Psi_n = \xi_n + \theta_n.$$ 

%and
% $$ \xi_n = 0.1sin(0.3n)+0.2sin(50n).$$

The algorithm's step-size or gain $a_{n}$ was scheduled as follows:
$$a_n = \frac{g}{(1+n)^{\rho}},$$ where $g=0.08, \rho=0.95$, were observed to lead to quick convergence in under a hundred iterations using qSGD $\# 1$. 

% Hence, the structure of our exploration signal takes the form:

% \begin{gather*}\label{exploration}
% \begin{split} 
%  \xi_n = \begin{bmatrix}
%                 a^1sin(2\pi \omega^1 n)\\
%                 a^2sin(2\pi \omega^2 n)
%                 \end{bmatrix}
% \end{split}
% \end{gather*}
%and $a^1=0.1,a^2=0.2$. 

\section{Uniform vs. Partitioned Parameterized Policy Approach}
The partitioned approach is a generalized version of the uniform approach that learns a parameter for different regions in the state space whereas the uniform parameterized approach applies qSGD to learn the parameter and use that same parameter throughout the entire state-space. While there may be several different ways to train a partitioned policy, the partitioned policy approach in this paper is trained by learning a parameter for each region starting from some number of different initial conditions in each region. In this paper only four regions are considered to make the point that better reductions in the cost-to-go function can be achieved when using a partitioned approach rather than a uniform approach, trained with qSGD $\# 1$. See the diagram below to see how the parameters fit within an abstract state space:

\begin{figure}[hbt!]
  \centering
{\includegraphics[height=6cm, width=10cm]{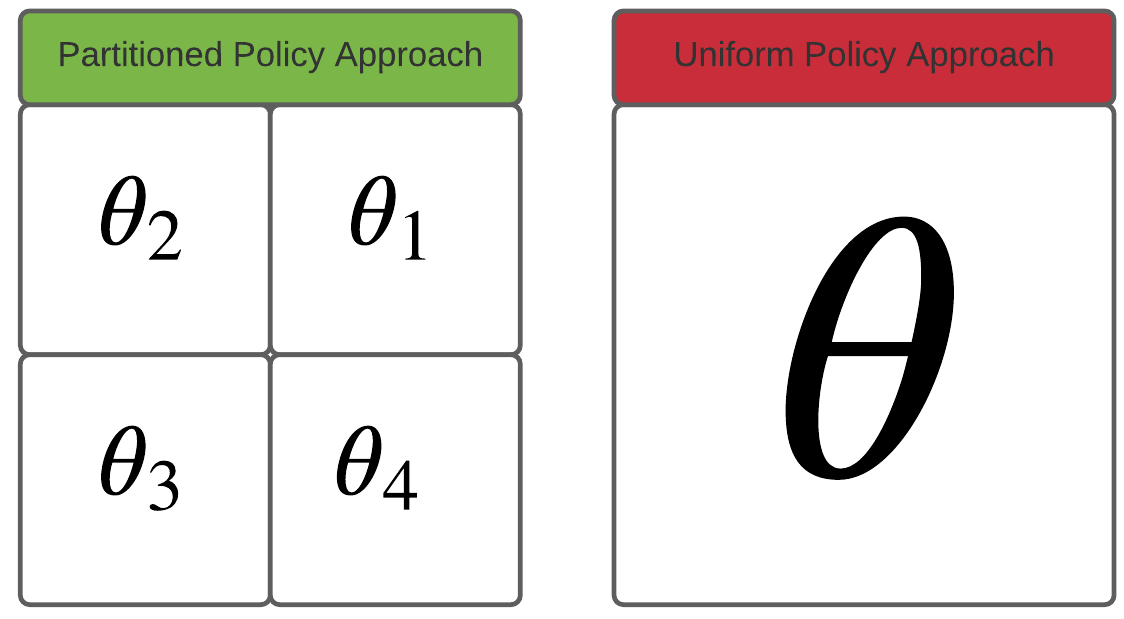}}
    \caption{Diagram of partitioned vs. uniform parameters covering an abstract state space}
  \label{fig:policy_diagram}
\end{figure}

The partitioned policy approach can be extended further to have as many divisions of the state space, and thus have as many parameters as are required to obtain a desired performance. This approach offers much more flexibility for parameterized control over the state space by learning different parameters specific to the geometry of each region. It is shown in the next section that the additional parameters, one for each unique region, more effectively lowers the cost than a uniform parameter approach and leads to better generalizations of the policy. This approach is also much more efficient than a neural network approach, where millions of extra parameters are often introduced without much justification. The following plot, \autoref{fig:training_ic}, shows an example of training initial conditions, and illustrates the fact that the parameter for each region is trained using qSGD $\# 1$ based only on the initial conditions for that region. For all experiments, $m=80$ initial conditions were used for training the uniform policy approach with the qSGD $\# 1$ algorithm, and the partitioned approach divides that up to $20$ training initial conditions per region. The number of iterations to run the algorithm was $N=50$ and $T_{max}=500$ for the simulation of the current iteration's policy parameters. In the next section, I show some closed-loop behavior and histograms to compare the performance of the different parameterized policy approaches.

Training Initial Conditions:
\begin{figure}[hbt!]
  \centering
{\includegraphics[height=6cm, width=10cm]{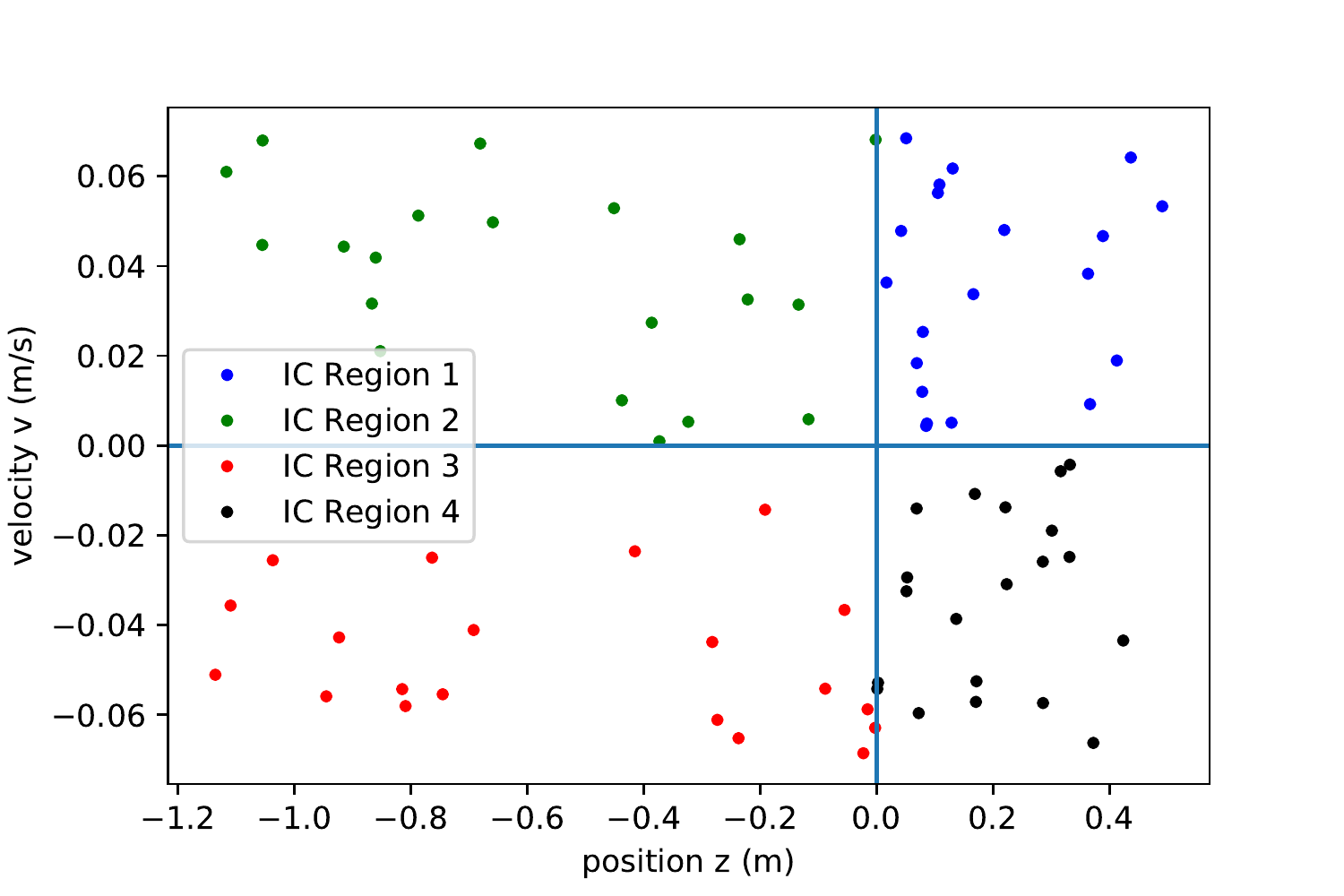}}
    \caption{Example training initial conditions for the partitioned and uniform policy approaches}
  \label{fig:training_ic}
\end{figure}

\section{Simulation Results}
The first plot of this section, \autoref{fig:uniform_vs_random_cl}, compares a randomly selected policy parameter and a convergent optimized uniform parameter on a random test initial condition:

\begin{figure}[hbt!]
  \centering
{\includegraphics[height=5cm, width=10cm]{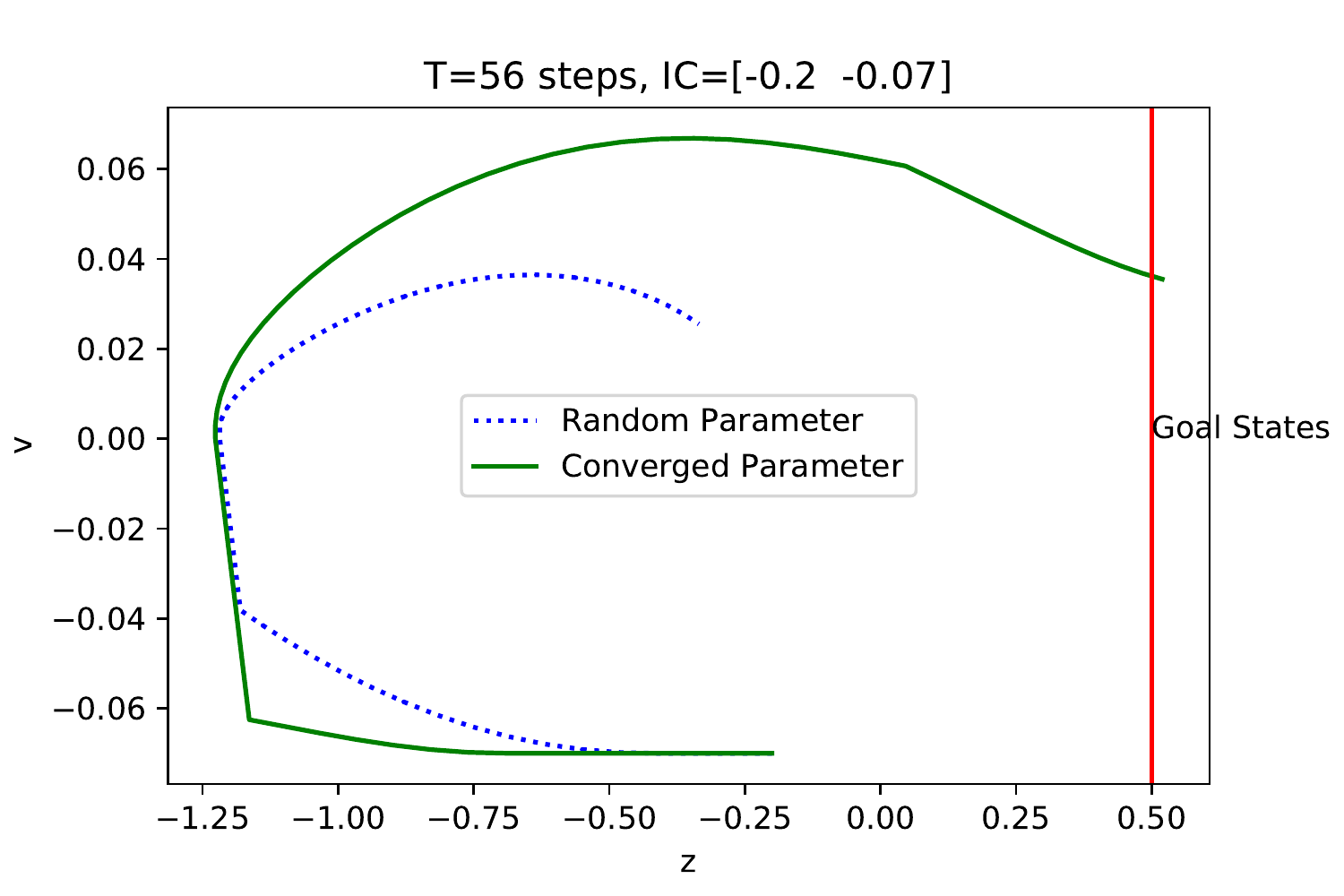}}
    \caption{Uniform vs. Random parameterized policy closed-loop behavior}
  \label{fig:uniform_vs_random_cl}
\end{figure}

\begin{figure}[hbt!]
  \centering
{\includegraphics[height=5cm, width=10cm]{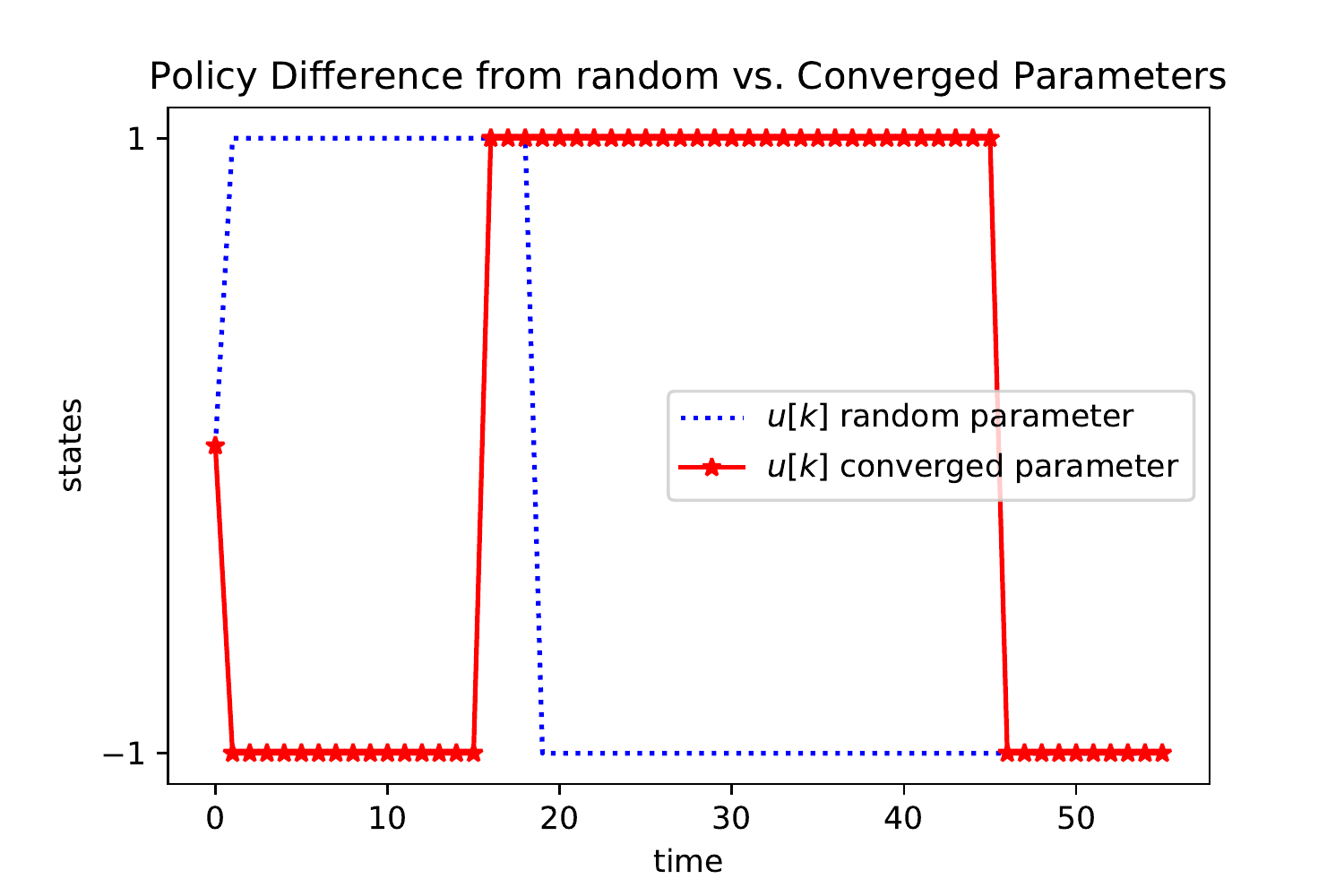}}
    \caption{Sequence of Control Decisions for Uniform vs. Random Approach}
  \label{fig:uniform_control_seq}
\end{figure}

\begin{figure}[hbt!]
  \centering
{\includegraphics[height=5cm, width=10cm]{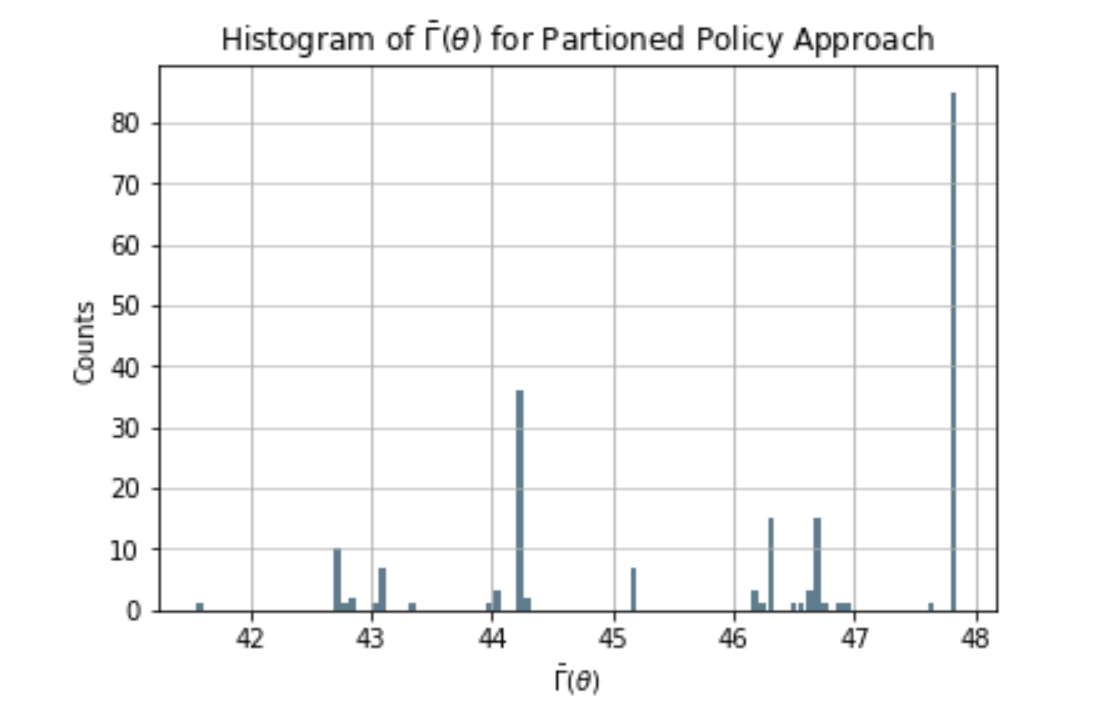}}
    \caption{Partitioned Control Policy Performance Histogram}
  \label{fig:partitioned_histogram}
\end{figure}

From this figure, it is clear that the uniform parameterized policy has learned that it is better to race to leftmost wall as quickly as possible to reset its velocity to zero after reaching the farthest position to the left in the environment; The car then darts across to the goal state. The next plot, \autoref{fig:uniform_control_seq} shows the sequence of control decisions that were made by these parameterized policies over the course of $T=56$ steps before the goal state was eventually reached using the uniform parameterized approach. Before showing anymore closed-loop behavior for the different approaches, I first show how more effective the partitioned approach is over the uniform approach by examining the histograms of the final $\Gamma(\theta)$, based on the $\theta$ at the end of the run, after convergence. These histograms, \autoref{fig:uniform_histogram} and \autoref{fig:partitioned_histogram}, quantify how much less cost the partitioned approach incurs over the uniform approach. To compare scalar performances, the final performance measure for the partitioned approach takes the final $\{\Gamma^i(\theta): 1\leq i \leq 4\}$ and averages them to obtain the average minimum time metric $\Bar{\Gamma}(\theta) = \frac{1}{4}\sum\limits_{i=1}^4\Gamma^i(\theta)$. The final $\Gamma(\theta)$ values based on the converged $\theta$ parameters are averaged over each region to arrive at an average minimum time for the partitioned approach $\Bar{\Gamma}(\theta)$. A histogram of this average performance is compared against the minimum time performance measure mentioned earlier for the uniform parameterized policy approach. 

% \begin{figure}[hbt!]
%   \centering
% {\includegraphics[height=6cm, width=10cm]{uniform_control_region.png}}
%     \caption{Uniform Control Policy for Entire State Space}
%   \label{fig:uniform_control_region}
% \end{figure}

% The third plot better illustrates the control decision that will be executed based on what state the mountain car is in. This plot confirms my observation of the mountain car's policy when it is in the farthest lowest left region of the state space. The next plots showcase closed-loop behavior for the partitioned policy approach to see how the policy changes based on the region in which the current state is active. 

The first plot is the uniform policy's performance histogram:

\begin{figure}[hbt!]
  \centering
{\includegraphics[height=5cm, width=10cm]{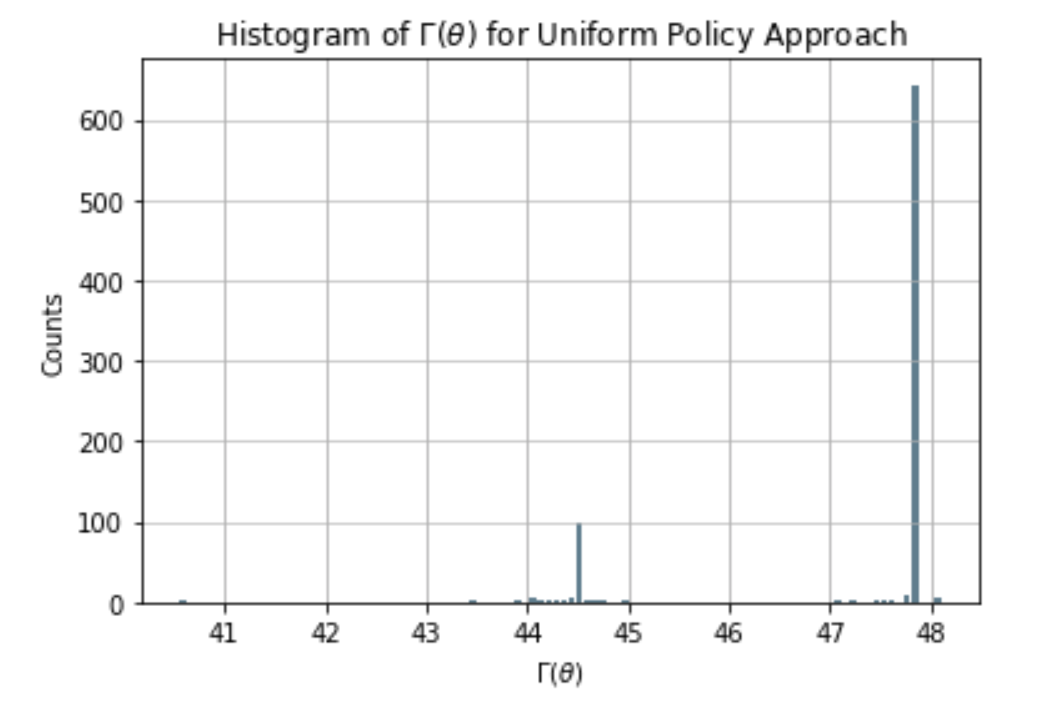}}
    \caption{Uniform Control Policy Performance Histogram}
  \label{fig:uniform_histogram}
\end{figure}

Next, the Partitioned policy's performance histogram is shown \autoref{fig:partitioned_histogram}.

\begin{figure}[hbt!]
  \centering
{\includegraphics[height=6cm, width=9cm]{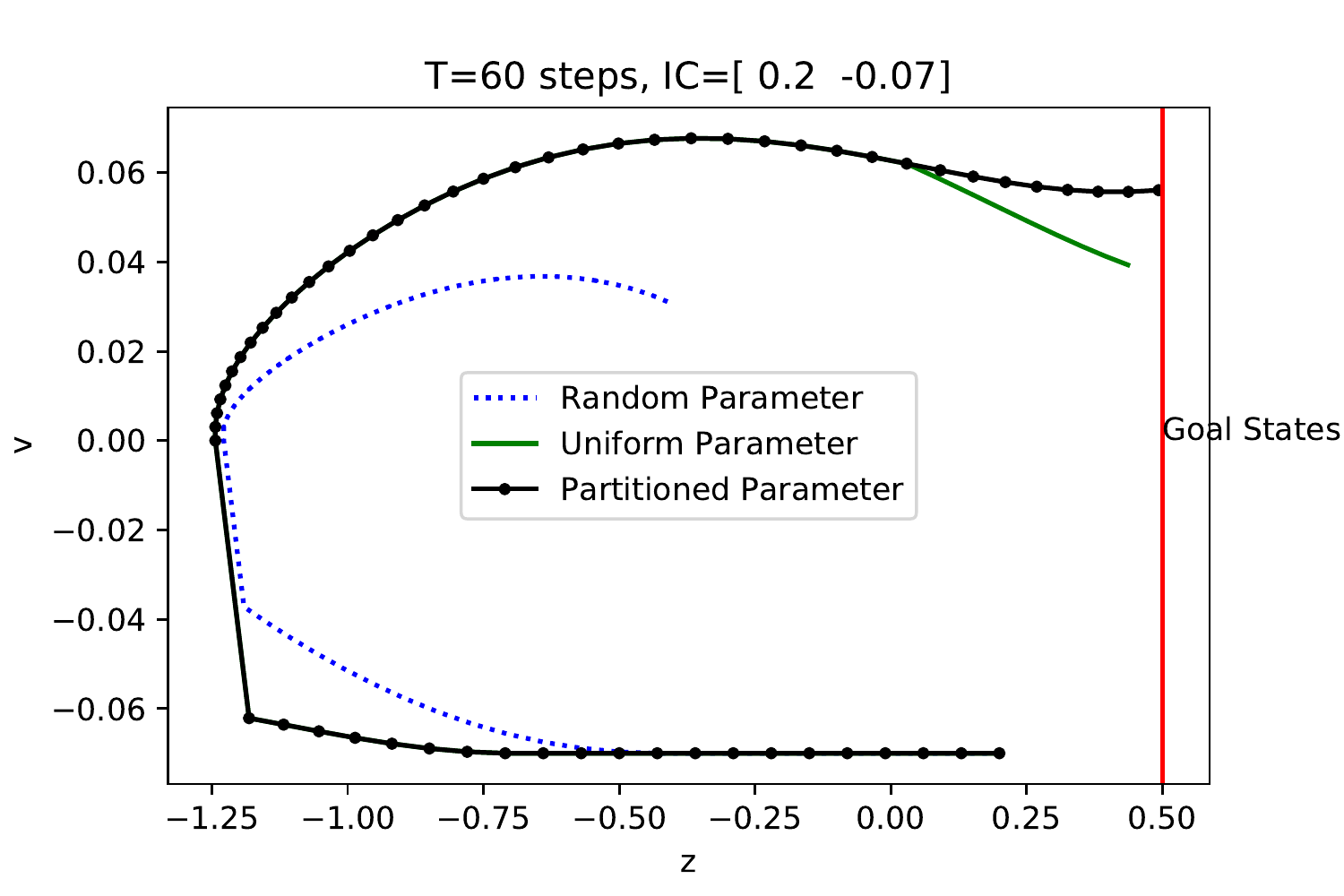}}
    \caption{Test closed loop behavior 1}
  \label{fig:test_cl_1}
\end{figure}

\begin{figure}[hbt!]
  \centering
{\includegraphics[height=6cm, width=9cm]{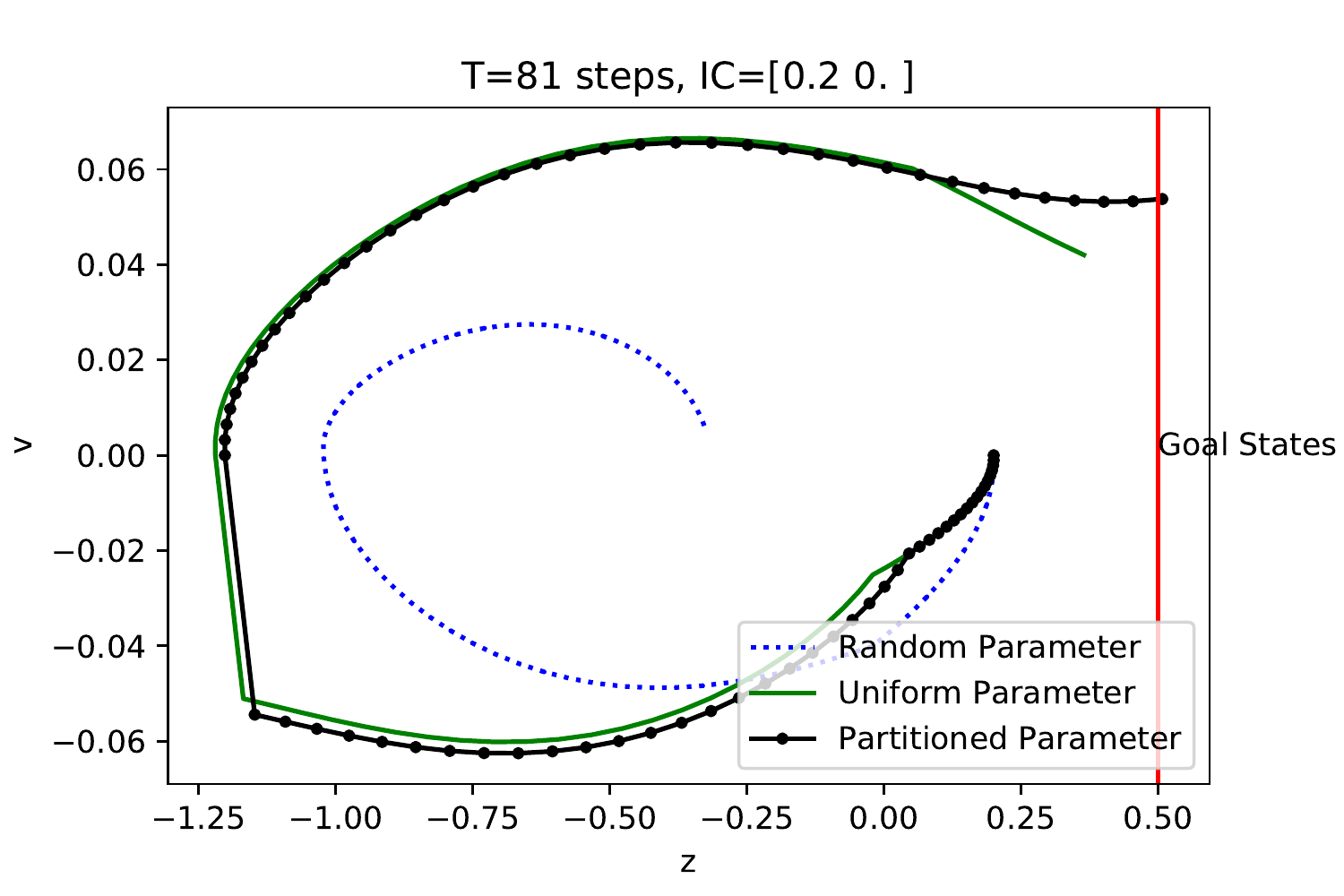}}
    \caption{Test closed loop behavior 2}
  \label{fig:test_cl_2}
\end{figure}

Using the partitioned approach consistently led to better closed-loop performance and a better overall performance in terms of reaching the top of the hill in attempting to minimize the minimum time objective. Looking at these histograms it is clear there are more modes for the partitioned approach at lower cost values around $T=43,~T=44,~T\in[46,47],$ and a large amount between $T\in[47,48]$. Where as the uniform approach predominately has the large cluster in the range $T\in[47,48]$ and a smaller but non-negligible cluster in $T\in[44,45]$. Note that the uniform histogram was computed from $M=800$ different initial parameters, whereas the partitioned approach used $800$ parameters but to make a fair comparison based on how its policy is implemented divides them up equally into $200$ initial parameters per region. Next, comparisons are made between the closed-loop performance of the uniform vs. partitioned policies, under non-ideal test initial conditions that were not a part of the training of these policies \autoref{fig:test_cl_1}, \autoref{fig:test_cl_2}.

% Based on the best found training parameters for the partitioned approach, the next plot illustrates how the control decisions vary across the state space. The control decision region for the partitioned approach learns a rounded curve boundary in the phase space. At first glance it is not clear what this division of this policy over the state space achieves, but the partitioned approach is undoubtedly more reliable and better performing than the uniform approach based on the histogram's count of performances.

% \begin{figure}[hbt!]
%   \centering
% {\includegraphics[height=6cm, width=10cm]{partitioned_region.png}}
%     \caption{Partitioned Control Policy for Entire State Space}
%   \label{fig:partitioned_policy}
% \end{figure}

\begin{figure}[hbt!]
  \centering
{\includegraphics[height=5cm, width=9cm]{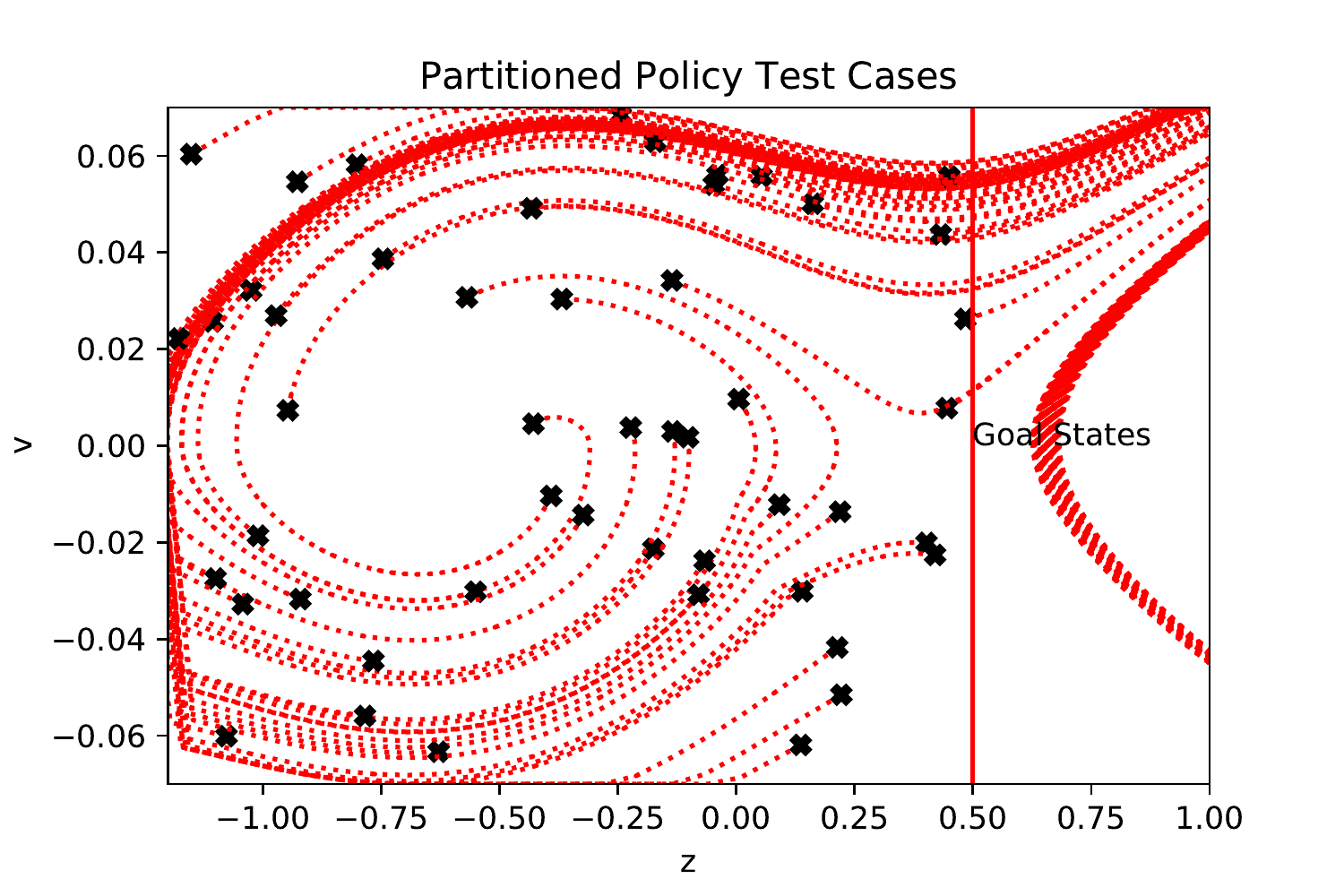}}
    \caption{Mountain Car's phase space trajectories using the partitioned parameterized policy}
  \label{fig:test_pp}
\end{figure}

\begin{figure}[hbt!]
  \centering
{\includegraphics[height=5cm, width=9cm]{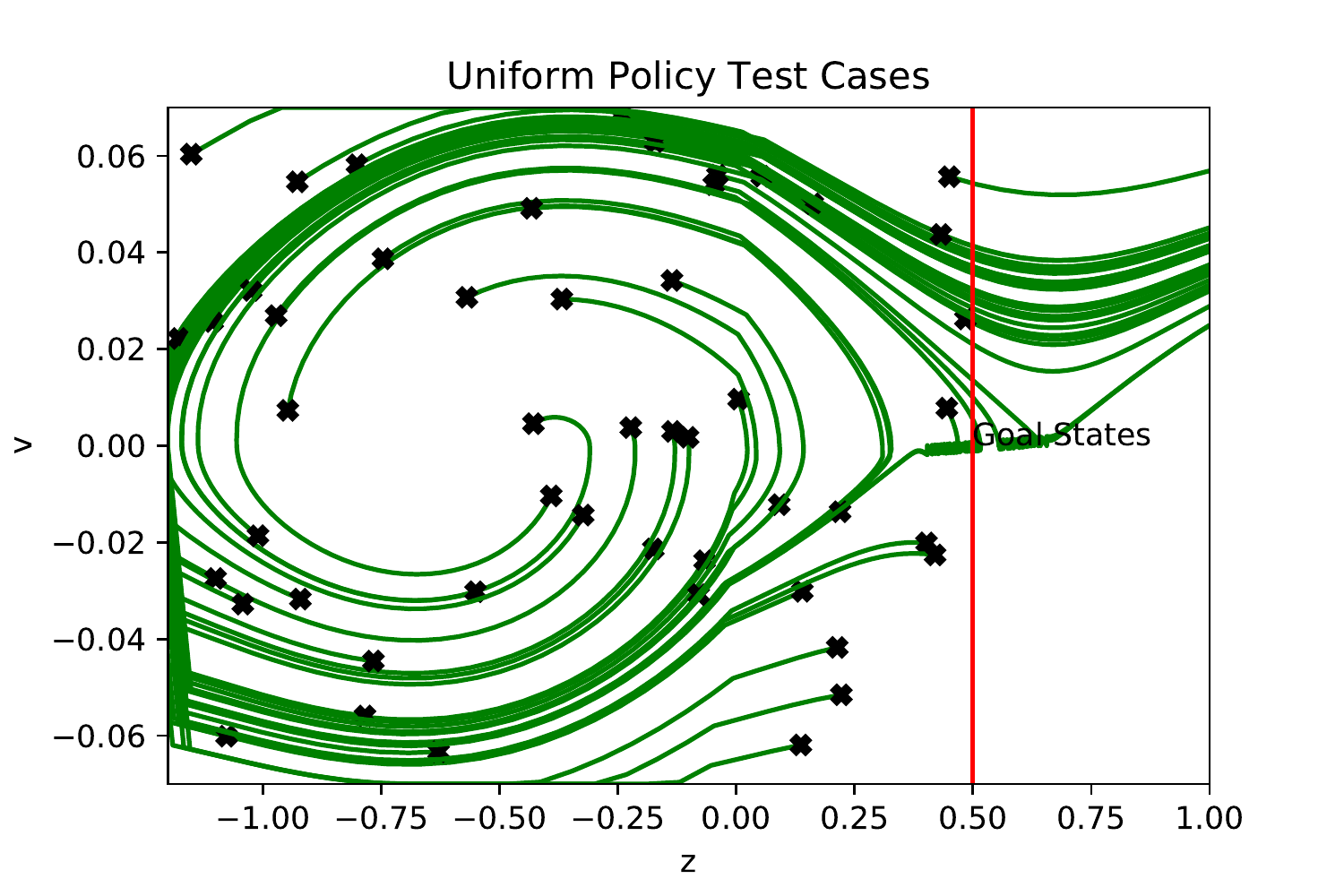}}
    \caption{Mountain Car's phase space trajectories using the uniform parameterized policy}
  \label{fig:test_up}
\end{figure}

The final two plots, \autoref{fig:test_pp} and \autoref{fig:test_up}, illustrate the generalizing ability of the partitioned and uniform policies to the entire state-space by testing under the same $50$ random initial conditions, after having been trained using qSGD $\# 1$. The black \say{X}s in both plots represent the starting test initial condition. Both the uniform and partitioned policy approaches converge to the goal state from every single test initial condition. Nevertheless, upon closer inspection of the first quadrant in the uniform approach plot \autoref{fig:test_up} reveals it is less efficient for several of those test initial conditions, and as a result, the Mountain Car has to travel all the way across the state space and back again to the goal state. Meanwhile, for those same test initial conditions in the first quadrant, the partitioned approach allows the Mountain Car to jet to the goal state almost in a straight shot. Some of the test initial conditions are also much more efficient from the second quadrant as well. 

% The closed-loop behavior plots  illustrate the varying levels of generalizability of the different policy approaches, with the partitioned approaches winning all of the convergent cases. However, during testing it was discovered that there are some test initial conditions for which all of the parameterized policies become trapped in cyclical dynamics for which the state trajectory cannot escape and head towards the goal state. See below \autoref{fig:test_failure_1}, \autoref{fig:test_failure_2}:

% \begin{figure}[hbt!]
% %\label{fig:testfail1}
%   \centering
% {\includegraphics[height=6cm, width=9cm]{test_failure_1.png}}
%     \caption{test failure 1}
%   \label{fig:test_failure_1}
% \end{figure}

% \begin{figure}[hbt!]
% %\label{fig:testfail2}
%   \centering
% {\includegraphics[height=6cm, width=9cm]{test_failure_2.png}}
%     \caption{test failure 2}
%   \label{fig:test_failure_2}
% \end{figure}

\section{Conclusion and Future Work}
Based on the simulation results, the training of the nonlinear parameterized feedback policies with qSGD $\#1$ appears to have solved the main issue of generalization in the Mountain Car problem, that was unresolved and noted in \cite{sutton2018reinforcement} about the Mountain Car being trapped in circular trajectories. From the experiments involving the many test initial conditions, it is clear that the qSGD $\# 1$ algorithm has produced robust policies for previously unseen starting points and most likely will lead to generalization success in many other continuous state control problems. Upon examining \autoref{fig:test_pp} and \autoref{fig:test_up} more closely, one can see that for some starting initial conditions in the upper right quadrant of the state space, the partitioned policy approach is able to leverage its different policy parameters in different regions to propel the Mountain Car almost immediately to the goal state, whereas the uniform approach does not have the correct parameter for that geometry and takes a full cycle to the leftmost wall before coming all the way back across the environment to make it back to the goal state. Similar efficiency gains are found in quadrant two thanks to the extra parameters offered by the partitioned parameterized policy approach. Despite the inefficiencies of the uniform approach, it still led to convergence in all cases after its parameter was trained with qSGD $\#1$. Furthermore, the partitioned approach based on the performance histograms led to consistently quicker runs as evidenced by having modes with lower costs. Future research should consider dividing the state-space further to see if more performance can be gained and if other ways to train the partitioned approach can be created. It will also be interesting to see if these methods can scale up to higher dimension continuous state control problems or be applied with success in other continuous space RL tasks. 
\newpage

\bibliographystyle{unsrt}
\bibliography{main}

\begin{thebibliography}{10}

\bibitem{sutton2018reinforcement}
Richard~S Sutton and Andrew~G Barto.
\newblock {\em Reinforcement learning: An introduction}.
\newblock MIT press, 2018.

\bibitem{moore1990efficient}
Andrew~William Moore.
\newblock Efficient memory-based learning for robot control.
\newblock 1990.

\bibitem{suttongeneralization}
Richard~S Sutton.
\newblock Generalization in reinforcement learning: Successful examples using
  sparse coarse coding.

\bibitem{singh1996reinforcement}
Satinder~P Singh and Richard~S Sutton.
\newblock Reinforcement learning with replacing eligibility traces.
\newblock {\em Machine learning}, 22(1):123--158, 1996.

\bibitem{heidrich2008variable}
Verena Heidrich-Meisner and Christian Igel.
\newblock Variable metric reinforcement learning methods applied to the noisy
  mountain car problem.
\newblock {\em Recent Advances in Reinforcement Learning}, page 136, 2008.

\bibitem{smartpractical}
William~D Smart and Leslie~Pack Kaelbling.
\newblock Practical reinforcement learning in continuous spaces.

\bibitem{melnikov2014projective}
Alexey~A Melnikov, Adi Makmal, and Hans~J Briegel.
\newblock Projective simulation applied to the grid-world and the mountain-car
  problem.
\newblock {\em arXiv preprint arXiv:1405.5459}, 2014.

\bibitem{CSRL}
Sean Meyn.
\newblock {\em Control Systems and Reinforcement Learning}.
\newblock {Cambridge University Press} (In preparation), 2021.

\bibitem{Meyn2000ode}
Vivek~S Borkar and Sean~P Meyn.
\newblock The ode method for convergence of stochastic approximation and
  reinforcement learning.
\newblock {\em SIAM Journal on Control and Optimization}, 38(2):447--469, 2000.

\bibitem{bowyer2021predictor}
Caleb Bowyer.
\newblock Predictor-corrector (pc) temporal difference (td) learning (pctd).
\newblock {\em arXiv preprint arXiv:2104.09620}, 2021.

\bibitem{Devraj}
AM~Devraj, A~Bu{\v{s}}ic, and S~Meyn.
\newblock Fundamental design principles for reinforcement learning algorithms.
\newblock {\em Handbook on Reinforcement Learning and Control. Springer}, 2020.

\end{thebibliography}

% \bibitem[Moore, Andrew William]
% \textit{Efficient memory-based learning for robot control} 
% 1990 Citeseer

% References follow the acknowledgments. Use unnumbered first-level heading for
% the references. Any choice of citation style is acceptable as long as you are
% consistent.

% It is permissible to reduce the font size to \verb+small+ (9 point)
% when listing the references.
% Note that the Reference section does not count towards the page limit.
% \medskip

% {
% \small

% [1] Alexander, J.A.\ \& Mozer, M.C.\ (1995) Template-based algorithms for
% connectionist rule extraction. In G.\ Tesauro, D.S.\ Touretzky and T.K.\ Leen
% (eds.), {\it Advances in Neural Information Processing Systems 7},
% pp.\ 609--616. Cambridge, MA: MIT Press.

% [2] Bower, J.M.\ \& Beeman, D.\ (1995) {\it The Book of GENESIS: Exploring
%   Realistic Neural Models with the GEneral NEural SImulation System.}  New York:
% TELOS/Springer--Verlag.

% [3] Hasselmo, M.E., Schnell, E.\ \& Barkai, E.\ (1995) Dynamics of learning and
% recall at excitatory recurrent synapses and cholinergic modulation in rat
% hippocampal region CA3. {\it Journal of Neuroscience} {\bf 15}(7):5249-5262.

%%%%%%%%%%%%%%%%%%%%%%%%%%%%%%%%%%%%%%%%%%%%%%%%%%%%%%%%%%%%
\section*{Checklist}

% %%% BEGIN INSTRUCTIONS %%%
% The checklist follows the references.  Please
% read the checklist guidelines carefully for information on how to answer these
% questions.  For each question, change the default \answerTODO{} to \answerYes{},
% \answerNo{}, or \answerNA{}.  You are strongly encouraged to include a {\bf
% justification to your answer}, either by referencing the appropriate section of
% your paper or providing a brief inline description.  For example:
% \begin{itemize}
%   \item Did you include the license to the code and datasets? \answerYes{See Section~\ref{gen_inst}.}
%   \item Did you include the license to the code and datasets? \answerNo{The code and the data are proprietary.}
%   \item Did you include the license to the code and datasets? \answerNA{}
% \end{itemize}
% Please do not modify the questions and only use the provided macros for your
% answers.  Note that the Checklist section does not count towards the page
% limit.  In your paper, please delete this instructions block and only keep the
% Checklist section heading above along with the questions/answers below.
% %%% END INSTRUCTIONS %%%

\begin{enumerate}

\item For all authors...
\begin{enumerate}
  \item Do the main claims made in the abstract and introduction accurately reflect the paper's contributions and scope?
    \answerYes{}
  \item Did you describe the limitations of your work?
    \answerYes{}
  \item Did you discuss any potential negative societal impacts of your work?
    \answerNo{}, As this research is more theoretical in nature, and is far removed from any sort of application. 
  \item Have you read the ethics review guidelines and ensured that your paper conforms to them?
    \answerYes{}
\end{enumerate}

\item If you are including theoretical results...
\begin{enumerate}
  \item Did you state the full set of assumptions of all theoretical results?
    \answerYes{}
	\item Did you include complete proofs of all theoretical results?
    \answerYes{}
\end{enumerate}

\item If you ran experiments...
\begin{enumerate}
  \item Did you include the code, data, and instructions needed to reproduce the main experimental results (either in the supplemental material or as a URL)?
    \answerYes{}, all of the code will be made available in jupyter-notebook form. The training and test initial conditions are saved in .npy format as numpy arrays so that in the future other kinds of policy methods could be compared against the results here, for reproducibility. 
  \item Did you specify all the training details (e.g., data splits, hyperparameters, how they were chosen)?
     \answerYes{}
	\item Did you report error bars (e.g., with respect to the random seed after running experiments multiple times)?
     \answerNo{}, I have run many repeated experiments using histograms, to ensure the results for each policy are statistically meaningful. 
	\item Did you include the total amount of compute and the type of resources used (e.g., type of GPUs, internal cluster, or cloud provider)?
    \answerYes{}, it is run on MacbookAir 1.1 GHz Quad-Core Intel Core i5. 
\end{enumerate}

\item If you are using existing assets (e.g., code, data, models) or curating/releasing new assets...
\begin{enumerate}
  \item If your work uses existing assets, did you cite the creators?
    \answerYes{}, the origins of the Mountain Car environment are summarized and credit given properly in the background and overview section.
  \item Did you mention the license of the assets?
    \answerNA{}
  \item Did you include any new assets either in the supplemental material or as a URL?
    \answerNA{}
  \item Did you discuss whether and how consent was obtained from people whose data you're using/curating?
    \answerNA{}
  \item Did you discuss whether the data you are using/curating contains personally identifiable information or offensive content?
    \answerNA{}
\end{enumerate}

\item If you used crowdsourcing or conducted research with human subjects...
\begin{enumerate}
  \item Did you include the full text of instructions given to participants and screenshots, if applicable?
    \answerNA{}
  \item Did you describe any potential participant risks, with links to Institutional Review Board (IRB) approvals, if applicable?
    \answerNA{}
  \item Did you include the estimated hourly wage paid to participants and the total amount spent on participant compensation?
    \answerNA{}
\end{enumerate}

\end{enumerate}

%%%%%%%%%%%%%%%%%%%%%%%%%%%%%%%%%%%%%%%%%%%%%%%%%%%%%%%%%%%%

% \appendix

% \section{Appendix}

% Optionally include extra information (complete proofs, additional experiments and plots) in the appendix.
% This section will often be part of the supplemental material.

\end{document}